%% file: main.tex
\documentclass[10pt,twocolumn,letterpaper]{article}

\usepackage{cvpr}
\usepackage{times}
\usepackage{epsfig}
\usepackage{graphicx}
\usepackage{amsmath}
\usepackage{amssymb}
\usepackage[normalem]{ulem}
\usepackage{afterpage}
\usepackage{booktabs}
\usepackage{color}
\usepackage{verbatim}
\usepackage{mathrsfs,amsmath}
\usepackage{soul}
\usepackage{xspace}
\usepackage{comment}
\usepackage{array}
\usepackage{multirow}
\usepackage{url}
\usepackage{epigraph}
\usepackage[toc,page]{appendix}
\usepackage{microtype}
\usepackage{pifont}
\usepackage{bm}

\usepackage[separate-uncertainty = true,multi-part-units = repeat]{siunitx}

\usepackage[pagebackref=true,breaklinks=true,letterpaper=true,colorlinks,bookmarks=false]{hyperref}

\cvprfinalcopy 

\def\cvprPaperID{1062} 

\ifcvprfinal\fi
\begin{document}

\title{4D Visualization of Dynamic Events from Unconstrained Multi-View Videos}
\author{Aayush Bansal \quad Minh Vo \quad Yaser Sheikh \quad Deva Ramanan \quad Srinivasa Narasimhan\\
Carnegie Mellon University\\
{\tt\small{\{aayushb,mpvo,yaser,deva,srinivas\}@cs.cmu.edu}}\\
\url{http://www.cs.cmu.edu/~aayushb/Open4D/}}

\maketitle

\input{abstract}

\input{introduction}

\input{background.tex}

\input{method.tex}

\input{multi-view-sequences}

\input{experiments.tex}

\input{discussion.tex}

{\small
\bibliographystyle{ieee_fullname}
\bibliography{main.bbl}
}

\end{document}

%% file: abstract.tex
\begin{abstract}
\vspace{-0.5cm}

We present a data-driven approach for 4D space-time visualization of dynamic events from videos captured by hand-held multiple cameras.  Key to our approach is the use of self-supervised neural networks specific to the scene to compose static and dynamic aspects of an event. Though captured from discrete viewpoints, this model enables us to move around the space-time of the event continuously. This model allows us to create virtual cameras that facilitate: (1) freezing the time and exploring views; (2) freezing a view and moving through time; and (3) simultaneously changing both time and view. We can also edit the videos and reveal occluded objects for a given view if it is visible in any of the other views. We validate our approach on challenging in-the-wild events captured using up to 15 mobile cameras.

\end{abstract}

%% file: introduction.tex
\section{Introduction}
\label{sec:intro}

\begin{figure*}[t]
\centering
\includegraphics[width=1.0\linewidth]{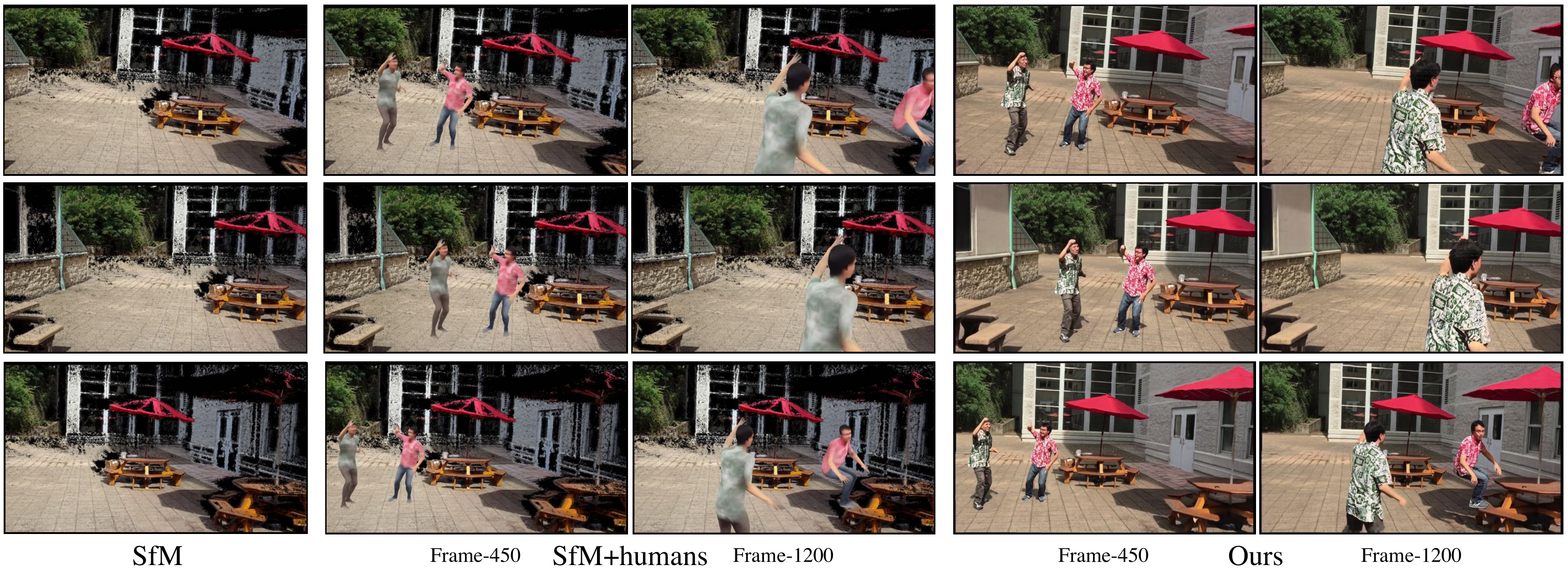}
\caption{\textbf{Comparison to existing work: } Given a dynamic event captured using $10$ phones, we \textbf{freeze time and explore views} for two time instances. We use a standard Structure-from-Motion (SfM)~\cite{schoenberger2016sfm, schoenberger2016mvs} to reconstruct the camera trajectory. As shown in first-column, SfM treats dynamic information as outliers for rigid reconstruction. We use additional cues such as 2D keypoints~\cite{cao2018openpose}, statistical human body model~\cite{loper2015smpl}, and human association~\cite{vo2018automatic} along-with the outputs of SfM to generate dynamic information for these two time instances (Frame-$450$ and Frame-$1200$ in second and third columns respectively). We call this \textbf{SfM+humans}. These three outputs lack \emph{realism}. Additionally, the reconstruction fails for non-Lambertian surfaces (see glass windows), non-textured regions (see umbrellas), and shadows (around humans). Our approach, on the other hand, can densely synthesize the various static and dynamic components, as shown in fourth and fifth columns for the same moments.}
\label{fig:concerns}
\end{figure*}

Imagine going back in time and revisiting crucial moments of your lives, such as your wedding ceremony, your graduation ceremony, or the first birthday of your child, immersively from any viewpoint. The prospect of building such a \textit{virtual time machine}~\cite{Reddy1999} has become increasingly realizable with the advent of affordable and high-quality smartphone cameras producing extensive collections of social video data. Unfortunately, people do not benefit from this broader set of captures of their social events. When looking back, we are likely to only look at one video or two when potentially hundreds might have been captured from different sources. We present a data-driven approach that leverages all perspectives to enable a more complete exploration of the event. With our approach, the benefits from each extra perspective that is captured leads to a more complete experience. We seek to automatically organize the disparate visual data into a comprehensive four-dimensional environment (3D space and time). The complete control of spatiotemporal aspects not only enables us to see a dynamic event from any perspective but also allows geometrically consistent content editing. This functionality unlocks many potential applications in the movie industry and consumer devices, especially as virtual reality headsets are becoming popular by the day. Figure~\ref{fig:teaser_fig} show examples of virtual camera views synthesized using our approach for an event captured from multi-view videos.

Prior work on virtualized reality~\cite{Joo_2017_TPAMI,kanade2006historical,kanade1996} has primarily been restricted to studio setups with tens or even hundreds of synchronized cameras. Four hundred hours of video data is uploaded on YouTube every minute. This feat has become possible because of the commercial success of high quality hand-held cameras such the iPhones or GoPros. Many public events are easily captured from multiple perspectives by different people. Despite this new form of big visual data, reconstructing and rendering the dynamic aspects have mostly been limited to studios and not for in-the-wild captures with hand-held cameras. Currently, there exists no method for fusing the information from multiple cameras into a single comprehensive model that could facilitate content sharing. This gap is largely because the mathematics of dynamic 3D reconstruction~\cite{hartley2003multiple} is not well-posed. The segmentation of objects~\cite{gupta2019lvis} are far from being consistently recovered to do 3D reconstruction~\cite{zhang1999shape}. Large scale analytics of internet images exist for static scenes~\cite{heinly2015_reconstructing_the_world,schoenberger2016sfm, schoenberger2016mvs,Snavely:2006} alone, and ignores the interesting dynamic events (as shown in Figure~\ref{fig:concerns}-first-column). 

We pose the problem of 4D visualization from in-the-wild captures within an image-based rendering paradigm utilizing large capacity parametric models. The parametric models based on convolutional neural nets (CNNs) can circumvent the requirement of explicitly computing a comprehensive model~\cite{UnstructuredVBR10,Carranza:2003} for modeling and fusing static and dynamic scene components. Key to our approach is the use of self-supervised CNNs specific to the scene to compose static and dynamic parts of the event. This data-driven model enables us to extract the nuances and details in a dynamic event. We work with in-the-wild dynamic events captured from multiple (not a fixed number)  mobile phone cameras. These multiple views have arbitrary baselines and unconstrained camera poses.
 
 \begin{figure*}[t]
\centering
\includegraphics[width=1.0\linewidth]{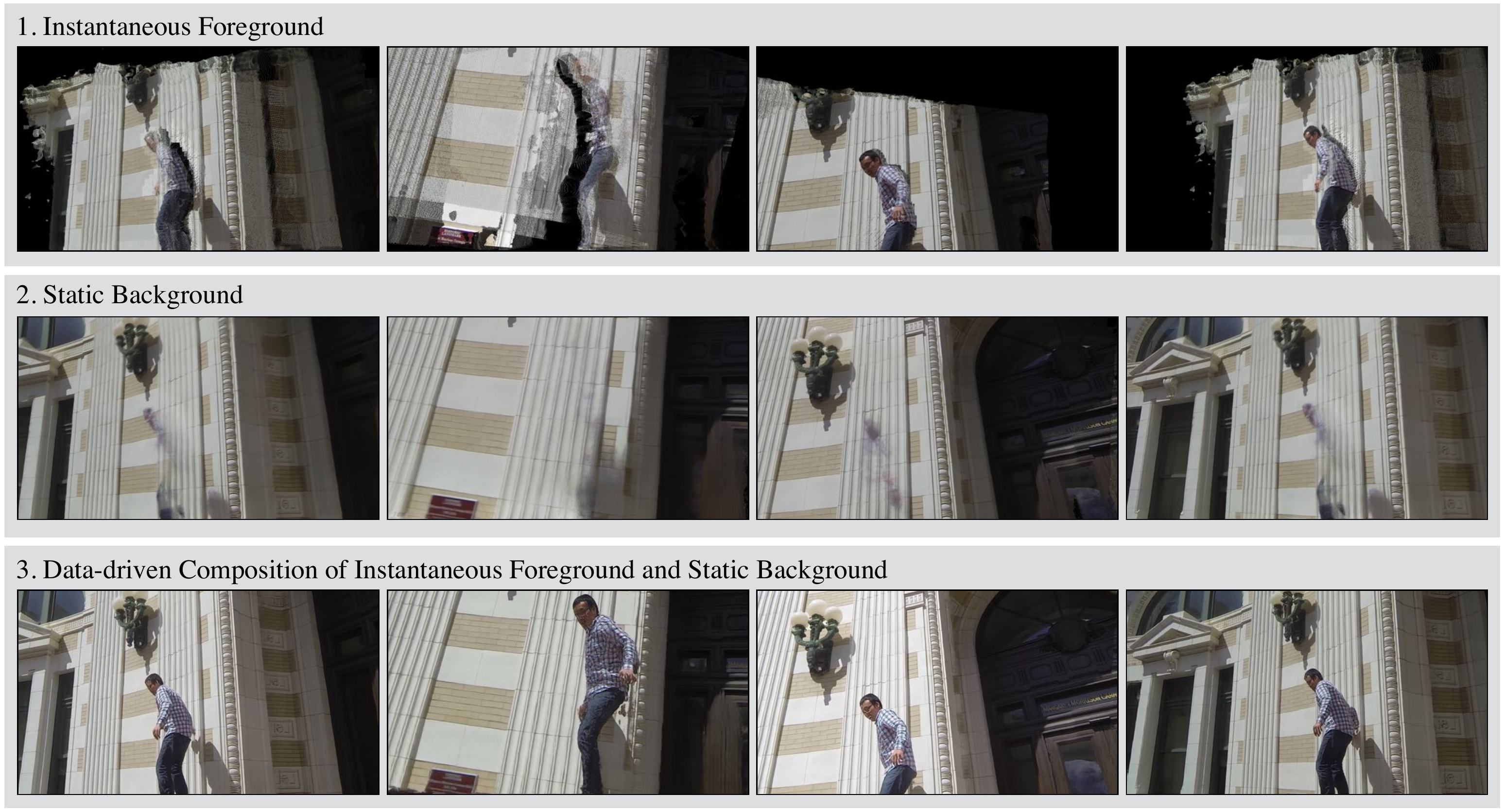}
\caption{\textbf{Overview: } We pose the problem of 4D visualization of dynamic events captured from multiple cameras as a data-driven composition of instantaneous foreground (\textbf{top}) and static background (\textbf{middle}) to generate the final output (\textbf{bottom}). The data-driven composition enables us to capture certain aspects that may otherwise be missing in the inputs, e.g., parts of the human body are missing in the first and second column, and parts of background are missing in first row.}
\label{fig:overview}
\end{figure*}

Despite impressive progress with CNN-based scene reconstruction~\cite{yao2018mvsnet, Huang2018, khot2019learning, Yang_2019_CVPR}, noticeable holes and artifacts are often visible, especially for large texture-less regions or non-Lambertian surfaces. We accumulate spatiotemporal information available from multiple videos to capture content that is not visible at a particular time instant. This accumulation helps us to capture even the large non-textured regions (umbrellas in Figure~\ref{fig:concerns}) or non-Lambertian surfaces (glass windows in Figure~\ref{fig:concerns}). Finally, a complete control of static and dynamic components of a scene, and viewpoint and time enables user-driven content editing in the videos. In public events, one often encounters random movement obstructing the cameras to capture an event. Traditionally nothing can be done about such spurious content in captured data. The complete 4D control in our system enables the user to remove unwanted occluders and obtain a clearer view of the actual event using multi-view information.

%% file: background.tex
\section{Related Work}
\label{sec:background}

There is a long history of 4D capture systems~\cite{kanade2006historical} to experience immersive virtualized reality~\cite{Fuchs-1994}, especially being able to see from any viewpoint that a viewer wants irrespective of the physical capture systems. 

\noindent\textbf{4D Capture in Studios: } The ability to capture depth maps from a small baseline stereo pair via 3D geometry techniques~\cite{hartley2003multiple} led to the development of video-rate stereo machines~\cite{kanade1996} mounting six cameras with small baselines. This ability to capture dense depth maps motivated a generation of researchers to develop close studios~\cite{Joo_2017_TPAMI,kanade1997virtualized,oswald2013convex,Zitnick:2004} that can precisely capture the dynamic events happening within it. A crucial requirement in these studios is the use of synchronized video cameras~\cite{kanade1997virtualized}. This line of research is restricted to a few places in the world with access to proper studios and camera systems. 

\noindent\textbf{Beyond Studios: } The onset of mobile phones have revolutionized the capture scenario. Each one of us possess high-definition smartphone cameras. Usually, there are more cameras at a place than there are people around. Many public events are captured by different people from various perspectives. This feat motivated researchers to use in-the-wild data for 3D reconstruction~\cite{heinly_dissertation,Snavely:2006} and 4D visualization~\cite{UnstructuredVBR10,Carranza:2003}. A hybrid of geometry~\cite{hartley2003multiple} and image-based rendering~\cite{shum2000review} approaches have been used to reconstruct 3D scenes from pictures~\cite{Debevec:1996}. Photo tourism~\cite{Snavely:2006} and the works following it~\cite{Agarwal2009,furukawa2010towards,furukawa2009accurate,heinly2015_reconstructing_the_world,sinha2008interactive} use internet-scale images to reconstruct architectural sites. These approaches have led to the development of immersive 3D visualization of static scenes.

The work on 3D reconstruction treats dynamic information as outliers and reconstructs the static components alone. Additional cues such as visual hulls~\cite{franco2005fusion,guillemaut2009robust,matusik2000image}, or 3D body scans~\cite{Carranza:2003,de2008performance}, or combination of both~\cite{ballan2008marker,vlasic2008articulated} are used to capture dynamic aspects (esp. human performances) from multi-view videos. Hasler et al.~\cite{hasler2009markerless} use markerless method by combining pose estimation and segmentation. Vedula et al.~\cite{Vedula-2005-9141} compute scene shape and scene flow for 4D modeling.  Ballan et al.~\cite{UnstructuredVBR10} model foreground subjects as video-sprites on billboards. However, these methods assume a single actor in multi-view videos. Recent approaches~\cite{Reddy_2018_CVPR,Vo_2016_CVPR} are not restricted by this assumption  but does sparse reconstruction. 

\begin{figure*}[t]
\centering
\includegraphics[width=1.0\linewidth]{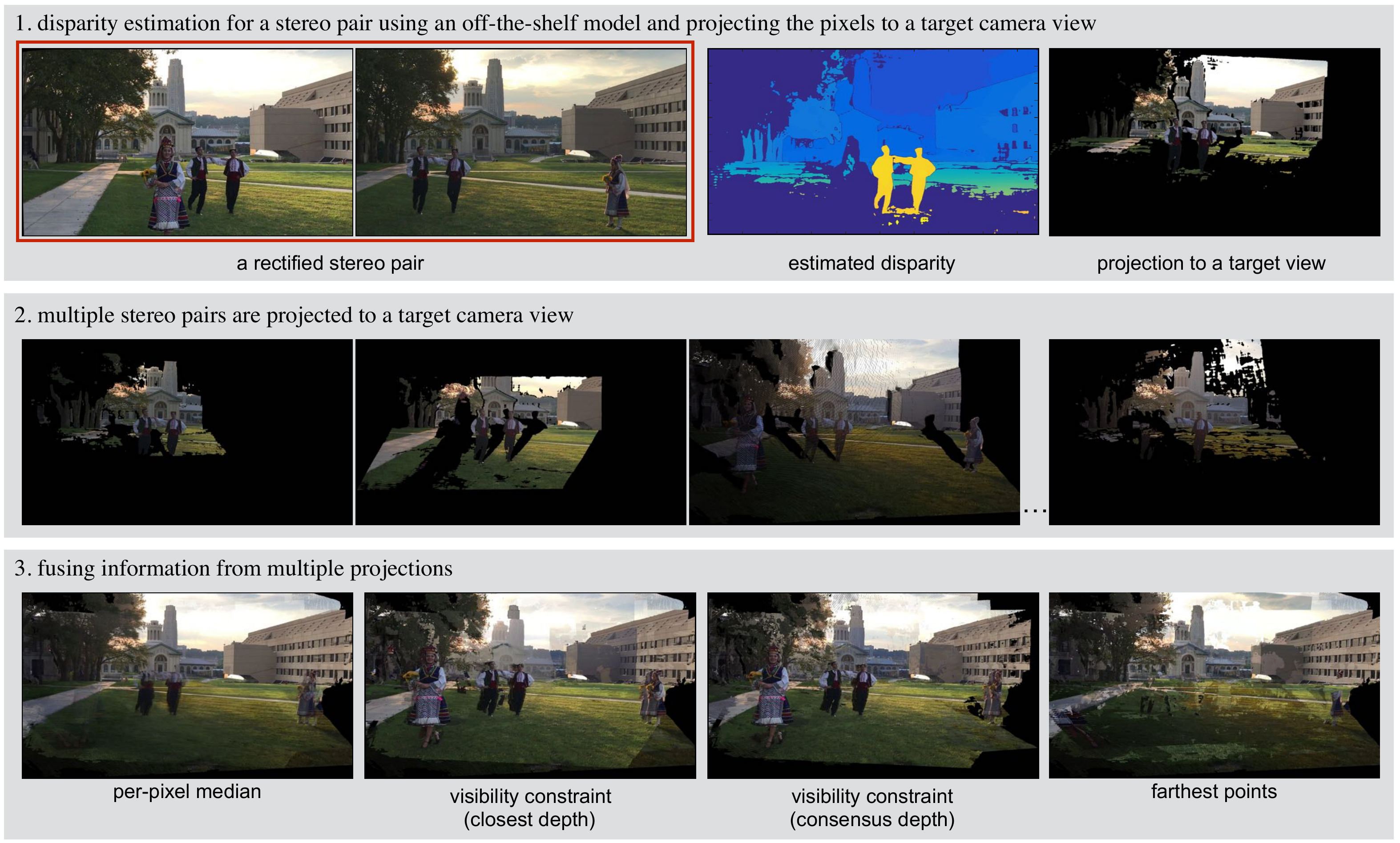}
\caption{\textbf{Instantaneous Foreground Estimation: } Given multiple stereo pairs and a target camera view, there are three steps for estimating instantaneous foreground. (1) We estimate disparity for a rectified stereo pair using an off-the-shelf disparity estimation approach~\cite{Yang_2019_CVPR} and project it to a target camera view using standard 3D geometry~\cite{hartley2003multiple}. (2) We repeat Step-1 for the $\binom{N}{2}$ stereo pairs. (3) A crucial aspect is to fuse the information from multiple projections to generate a comprehensive instantaneous foreground. A simple per-pixel median over the multiple projections leads to a loss of dynamic information because of the poor 3D estimates (shown in first column). We propose a visibility constraint using painter's algorithm. The visibility constraint builds a per-pixel cost volume of depth. It is natural to have multiple depth values for a given 2D pixel location in the image. For each pixel location, we consider the 3D point corresponding to the \emph{closest depth} to the target camera view. Due to noisy 3D estimates, it often leads to ghosting artifacts (shown in second column). However, we have multiple views and we  ensure consistency in the depth values of the closest 3D point for projection to a 2D pixel location by analyzing their visibility. As shown in third column, this \emph{consensus in depth} leads to improved results. Finally, we also show the projection of pixels corresponding to the \emph{farthest points} (fourth column) for a given pixel location to illustrate cost volume of depth. Note that it removes dynamic information in the scene.}
\label{fig:foreground}
\end{figure*}

\noindent\textbf{CNN-based Image Synthesis: } Data-driven approaches~\cite{denton2015deep,goodfellow2014generative,isola2017image,zhang2017stackgan} using convolutional neural networks~\cite{lecun2015deep} have led to impressive results in image synthesis. These results inspired a large body of work~\cite{Flynn_2019_CVPR,Flynn_2016_CVPR,Kalantari:2016,Meshry_2019_CVPR,srinivasan2019pushing,zhou2018stereo} on continuous view synthesis for small baseline shifts. Hedman et al.~\cite{Hedman:2018} extended this line of work to free-viewpoint capture. However, these methods are currently applicable to static scenes only. We combine the insights from CNN-based image synthesis and earlier work on 4D visualization to build a data-driven 4D Browsing Engine that makes minimal assumption about the content of multi-view videos.

%% file: method.tex
\section{4D Browsing Engine}
\label{sec:method}

We are given $N$ camera views with extrinsic parameters $\{C_1, C_2, ..., C_N\}$, and intrinsic parameters $\{M_1, M_2, ..., M_N\}$. Our goal is to generate virtual camera view $C$ that does not exist in any of these $N$ cameras. We temporally align all cameras using spatiotemporal bundle adjustment~\cite{Vo_2016_CVPR}, after which we assume the video streams are perfectly synchronized. Our method should be robust to possible alignment errors. Figure~\ref{fig:overview} shows an overview of our approach via a virtual camera that freezes time and explores views. We begin with estimation of instantaneous foreground in Section~\ref{ssn:fg}, static background in Section~\ref{ssn:bg}, and data-driven composition in Section~\ref{sec:ssc}. We finally discuss practical details and design decisions in Section~\ref{sec:cnn}.

\begin{figure*}[t]
\centering
\includegraphics[width=1.0\linewidth]{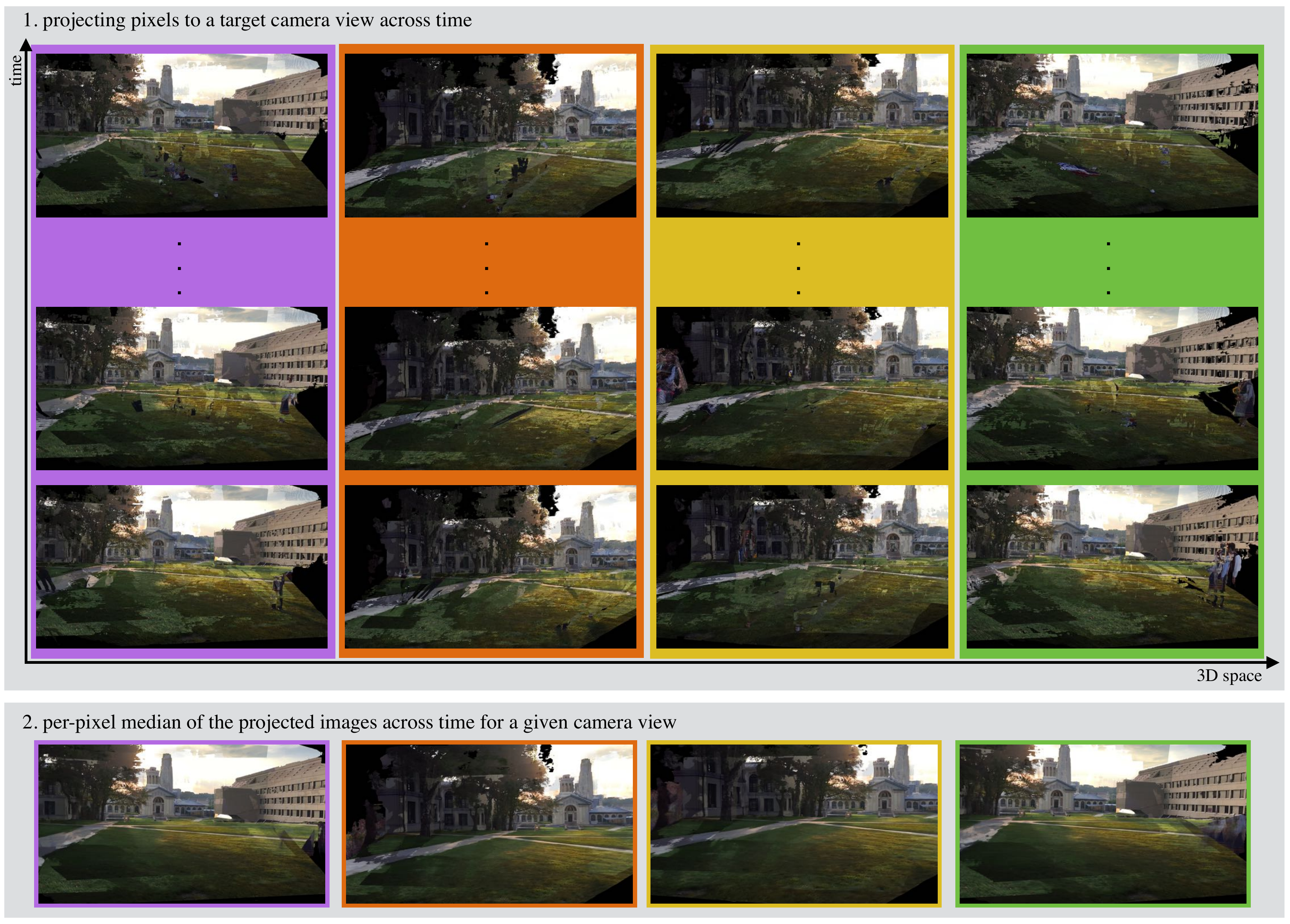}
\caption{\textbf{Static Background Estimation: } (1) We use the pixels corresponding to the farthest 3D points (from the per-pixel cost volume of depth) to generate an image for a target camera pose and time  instant. This enables us to get the farther components of the scene that are often stationary. However, this information is noisy. We, therefore compute the images for a given camera pose across all time. (2) A per-pixel median of the images over a large temporal window for a target camera pose results in a smooth stationary background.}
\label{fig:background}
\end{figure*}

\subsection{Instantaneous Foreground Estimation} 
\label{ssn:fg}

The $N$-camera setup provides us with $\binom{N}{2}$ stereo pairs. We build foreground estimates for a given camera pose and a time using multiple rectified stereo pairs. There are three essential steps for estimating instantaneous foreground (shown in Figure~\ref{fig:foreground}). (1) We begin with estimating disparity using an off-the-shelf disparity estimation module~\cite{Yang_2019_CVPR}. The knowledge of camera parameters for the stereo pair allows us to project the pixels to a target camera view using standard 3D geometry~\cite{hartley2003multiple}. Figure~\ref{fig:foreground}-1 shows an example of projection for a rectified stereo pair. (2) We repeat Step-1 for multiple stereo pairs. This step gives us multiple projections to a target camera view (shown in Figure-\ref{fig:foreground}-2) and per-pixel depth estimates. The projections from various $\binom{N}{2}$ stereo pairs tends to be noisy. This is due to sparse cameras, large stereo baseline, bad stereo-pairs, or errors in camera-poses. We cannot naively fuse the multiple projections to synthesize a target view in all conditions  for a target camera view. As an example, a simple per-pixel median on these multiple projection leads to the loss of dynamic information as shown Figure-\ref{fig:foreground}-3 (first column). (3) We enforce a visibility constraint to fuse information from multiple stereo pairs.

 \begin{figure*}[t]
\centering
\includegraphics[width=1.0\linewidth]{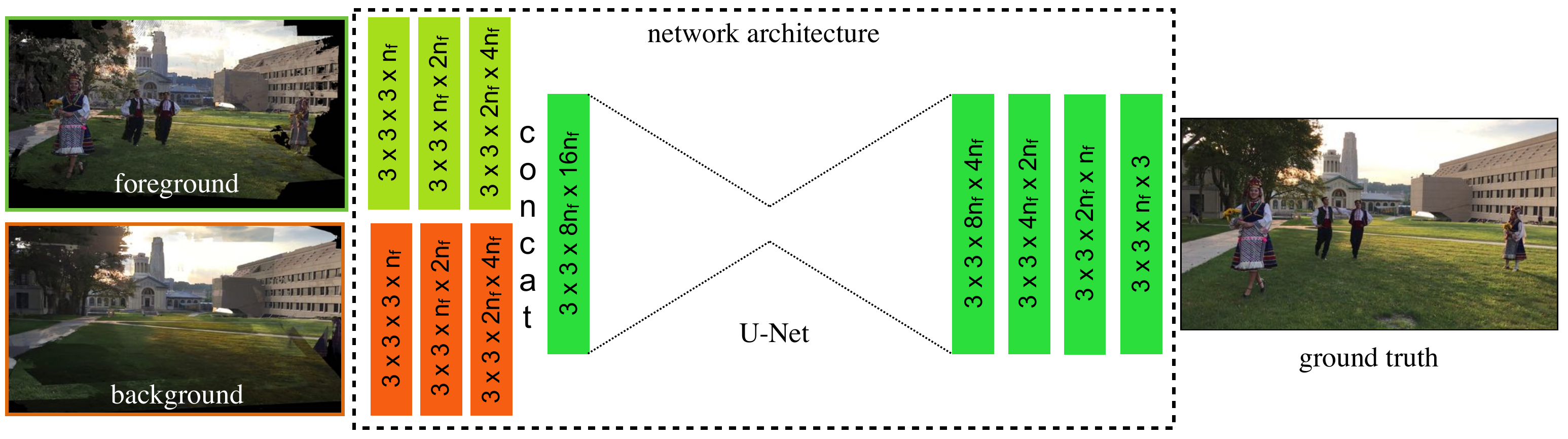}
\caption{\textbf{Self-Supervised Composition of Foreground and Background: } Given an input-output paired data created for a held out camera, we train a neural network to learn the composition. We show our neural network architecture that is used to compose the instantaneous foreground ($f_{c,t}$) and static background ($b_c$). The first part of our network consists of three convolutional layers. We concatenate the outputs of both foreground and background streams. The concatenated output is then fed forward to the standard U-Net architecture~\cite{ronneberger2015u}. Finally, the output of U-Net is then fed forward to three more convolutional layers that give the final image output.}
\label{fig:4d-network}
\end{figure*}

\noindent\textbf{Visibility Constraint: }  We use a visibility constraint that is a close approximation of painter's algorithm. The visibility constraint builds a per-pixel cost volume of depth that allows us to retrace the ray of light to its origin for a given camera view. Since we have multiple views, it is natural to have multiple depth values for a given 2D pixel location in the image. For each pixel location, we consider the 3D point corresponding to the closest depth. However, the closest 3D point from the cost volume of depth may not be accurate in our setup. As shown in the second column of Figure-\ref{fig:foreground}-3, we observe ghosting artifacts due to noisy 3D estimates. 

Fortunately, we can use multiple views to ensure consistency in the depth values of the closest 3D point that we should consider for projection at a given 2D pixel location. While a certain 3D location may not be viewed by all the cameras, it is highly likely that it would be seen in a significant subset of cameras. This means that we will estimate depth of same 3D point in our setup from many stereo pairs. We use this insight to build a consensus about the closest 3D point by ensuring that same depth value is observed in at least $3$ stereo pairs. We start with the smallest depth value ($>0$) at a pixel location in the cost volume and iterate till we find a consensus.  We ignore the pixel location (leave it as blank) if it is not observed from a sufficient number of cameras (less than 3). Ideally, the depth values of a 3D point (though computed from different stereo pairs) should be same from the target camera view. For the practical purposes, we define a threshold that can account for noise. As shown in third column of Figure-\ref{fig:foreground}-3, this leads to improved results. However, there is still missing information and ghosting artifacts in this output to qualify as a real image. 


\subsection{Static Background Estimation}
 \label{ssn:bg}
 
 The missing information in the output of  Section~\ref{ssn:fg} is due to the lack of visibility of points across multiple views for a given time instant. We accumulate long-term spatiotemporal information to compute static background for a target camera view.  The intrinsic and extrinsic parameters from $N$ physical cameras enable us to create the views over a large temporal window of $[0,t]$ for a target camera position. Figure~\ref{fig:background}-1 shows examples of virtual cameras for various poses and time instants. For a given camera pose and time instant, we once again use the dense per-pixel cost volume of depth. This time we use the farthest depth values to capture missing information (also refer to the last column of Figure~\ref{fig:foreground}-3). 
 
 The estimates corresponding to the farthest point for one time instant are noisy. We compute the images for a given camera pose across all time. A per-pixel median of different views (across time) for a given camera pose results in a smoother static background (albeit still noisy). Empirically, such median image computed over large temporal window for a given camera position also contain textureless and non-Lambertian stationary surfaces in a scene (observed consistently in Figure~\ref{fig:background}-2, Figure~\ref{fig:concerns}, and Figure~\ref{fig:overview}). We now have a pair of complementary signals: (1) sparse and noisy foreground estimates; and (2) dense but static background estimates.

\begin{figure*}[t]
\centering
\includegraphics[width=1.0\linewidth]{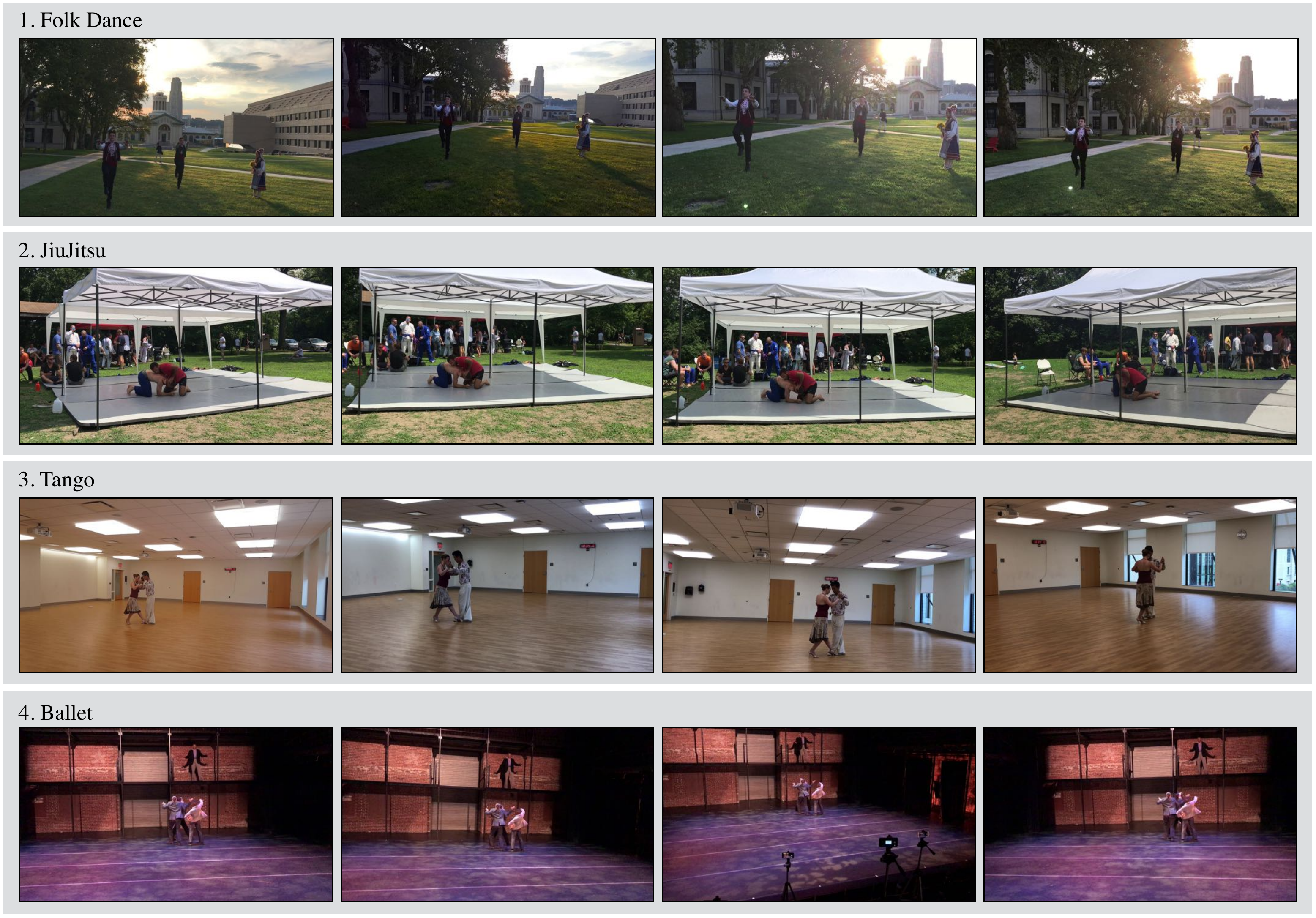} 
\caption{\textbf{Human Performances}: We captured a wide variety of human performances from multiple cameras. The human performances in these sequences have a wide variety of motion, clothing, human-human interaction, and human-object interaction . These sequences were captured in varying environmental and illumination condition. Shown here are examples from four such sequences to give a sense of extremely challenging setup dealing with a wide and arbitrary camera baseline.}
\label{fig:human_dataset}
\end{figure*}

\subsection{Self-Supervised Composition} 
\label{sec:ssc}

We use a data-driven approach to learn the fusion of instantaneous foreground, $F$, and static background, $B$, to generate the required target view for given camera parameters. The instantaneous foreground for a camera pose $c$ and time $t$ is notated as $f_{c,t}$. The static background for a camera pose $c$ is notated as $b_{c}$. However, there exists no ground truth or paired data to train such a model in a data-driven manner. We formulate the data-driven composition in a self-supervised manner by reconstructing a known held-out camera view, $o_{c,t}$, from the remaining $N-1$ views. This gives us a paired data $\{((f_{c,t},b_{c},), o_{c,t})\}$ for learning a mapping $G: (B, F) \rightarrow O$. We now have a pixel-to-pixel translation~\cite{isola2017image} and therefore, we can easily train a convolutional neural network (CNN) specific to a scene. We use three losses for optimization: (1) Reconstruction loss; (2) Adversarial loss; and (3) Frequency loss.

\noindent\textbf{Reconstruction Loss: }We use standard $l_1$ reconstruction loss to minimize reconstruction error on the content with paired data samples:

\begin{align}
  \min_{G} L_{r} =  \sum_{(c,t)} ||o_{c,t} - G(f_{c,t},b_{c})||_1
\label{eq:reg}
\end{align}

\noindent\textbf{Adversarial Loss: }  Recent work~\cite{goodfellow2014generative} has shown that learned mapping can be improved by tuning it with a discriminator $D$ that is adversarially trained to distinguish between real samples of $o_{c,t}$ from generated samples $G(f_{c,t}, b_c)$:

\begin{align}
\min_{G} \max_{D} L_{adv}(G,D) =  \sum_{(c,t)} \log D(o_{c,t})  + \nonumber\\ \sum_{(c,t)} \log (1 - D(G(f_{c,t},b_{c}))) 
\label{eq:adv}
\end{align}

\noindent\textbf{Frequency Loss: } We enforce a frequency-based loss function via Fast-Fourier Transform to learn appropriate frequency content and avoid generating spurious high-frequencies when ambiguities arise (inconsistent foreground and background inputs):

\begin{align}
  \min_{G} L_{fr} = \sum_{(c,t)} ||\mathscr{F}(o_{c,t}) - \mathscr{F}(G(f_{c,t},b_{c}))||_1
\label{eq:freq}
\end{align}

 where $\mathscr{F}$ is fast-Fourier transform. The overall optimization combines Eq.~\ref{eq:reg}, Eq.~\ref{eq:adv}, and Eq.~\ref{eq:freq}: 
 
 \begin{align}
  L = {\lambda_r}L_{r} + {\lambda_{adv}}L_{adv} + {\lambda_{fr}}L_{fr}\nonumber
\label{eq:overall}
\end{align}

where, $\lambda_r = \lambda_{fr} = 100$, and $\lambda_{adv} = 1$. Explicitly using background and foreground for target view makes the model independent of explicit camera parameters.

\begin{figure*}[t]
\centering
\includegraphics[width=1.0\linewidth]{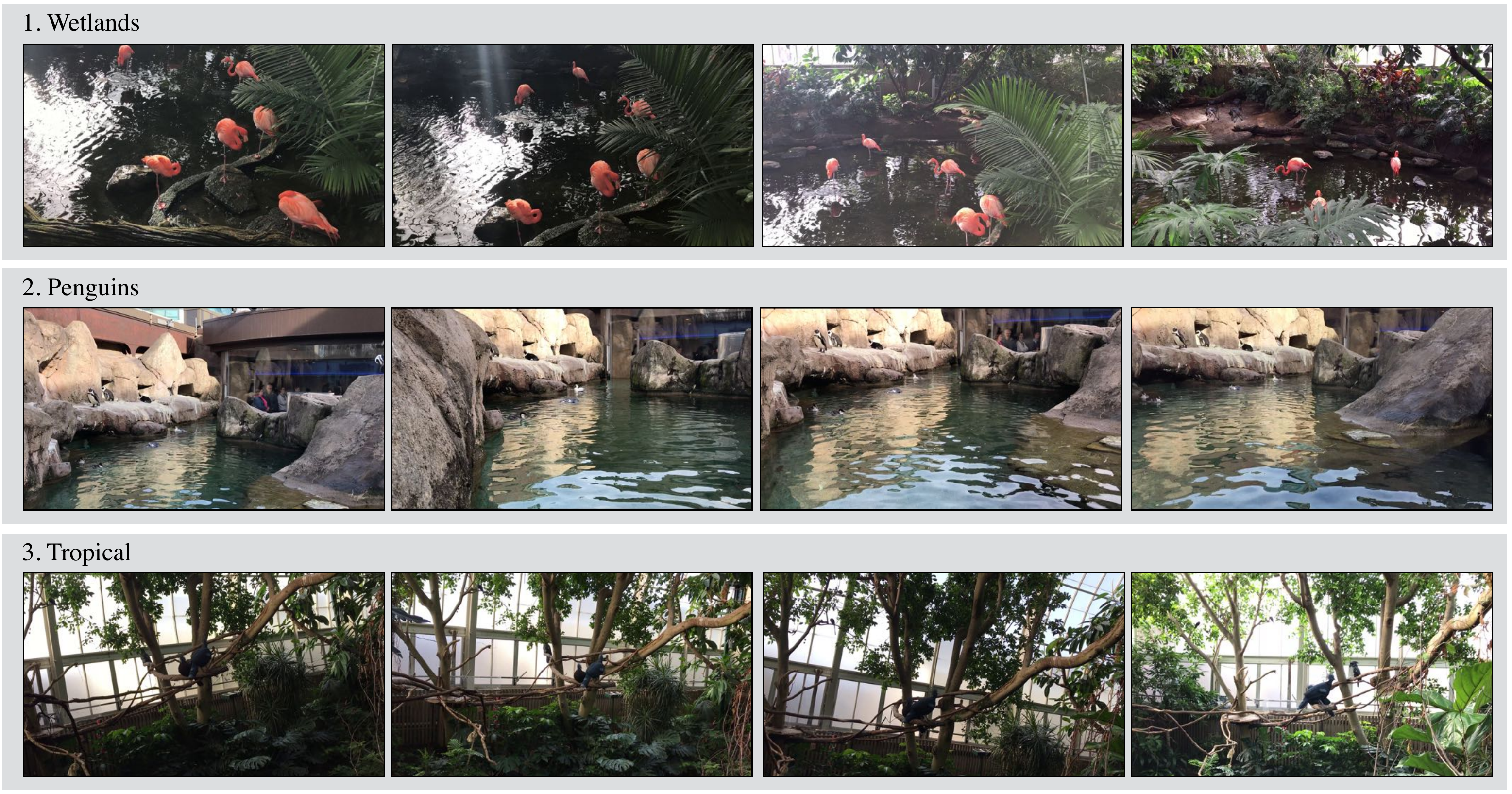} 
\caption{\textbf{Bird Sequences}: We captured a wide variety of birds at the National Aviary of Pittsburgh. We have no control on the motion of birds, their environment, lighting condition, and dynamism in the background due to human movement in the aviary. Shown here are examples from three such sequences to give a sense of our capture scenario.}
\label{fig:bird_dataset}
\end{figure*}

\noindent\textbf{ Network Architecture \& Optimization: }  We use HD-image inputs ($1080{\times}1920$). The input images are zero-padded making them $1280{\times}2048$ in dimensions. We use a modified U-Net architecture in our formulation as shown in Figure~\ref{fig:4d-network}. There are three parts of our neural network architecture. (1) The first part of our network consists of three convolutional layers. We concatenate the outputs of this part for both foreground and background stream. (2) The concatenated output is fed forward to the second part that is a standard U-Net architecture~\cite{ronneberger2015u}. (3) Finally, we feed the outputs of U-Net through three more convolutions that generates the final image. The batch-size is 1. The number of filter, $n_f$, in the first conv-layer of our network is $13$ (see Figure~\ref{fig:4d-network} for reference). For data augmentation purposes, we randomly resize the input to upto $1.5\times$ scale factor and randomly sample a crop from it. We use the Adam solver.  A model is trained from scratch with a learning rate of $0.0002$, and remains constant throughout the training. The model converges quickly in around $10$ epochs for a sequence.

\section{Practical Limitations and Design Decisions} 
\label{sec:cnn}

There are two practical challenges: (1) how to hallucinate missing information?; (2) how to avoid overfitting to specific input-output pair for a scene-specific CNN? In this section, we discuss the design decisions  that we made to overcome these challenges.

\subsection{Hallucinating Missing Information}
\label{ssn:hallucinating}

We have so far assumed the scenario where sufficient foreground and background information are available that can be composed to generate a comprehensive new view. However, it is possible that certain regions are never captured (e.g. blank pixels in foreground and background estimation in Figure~\ref{fig:foreground} and Figure~\ref{fig:background} respectively). This is possible due to many reasons such as sparse cameras, large stereo baseline, bad stereo-pairs, or errors in camera-poses. This is also possible if we try to visualize something beyond the convex hull of physical cameras. While filling smaller holes is reasonable for a parametric model, filling larger holes lead to temporally inconsistent artifacts. One way to deal with it is to learn a higher capacity model that may learn to fill larger holes. Other than fear of overfitting, the larger model needs prohibitive memory. Training a neural network combining the background and foreground information with HD images already require 30 GB memory of a v100 GPU.  In this work, we use a stacked multi-stage CNN to overcome this issue. 

\noindent\textbf{Multi-Stage CNN: } We use a high capacity model for low-res image generation that learns overall structure, and improve the resolution with multiple stages. We train three models for three different resolutions, namely: (1) low-res ($270{\times}480$); (2) mid-res ($540{\times}960$); and (3) hi-res ($1080{\times}1960$). These models are trained independently and form multiple stages of our formulation. At test time, we use these models sequentially starting from low-res to mid-res to hi-res outputs. This sequential processing is done when there are large holes in the instantaneous foreground and static background inputs. The output of the low-res model is used to fill the holes to the input of next model in sequence. Here we leverage the fact that we can have a very high capacity model for a low-resolution image and that holes become smaller as we make the image smaller. The overall network architecture remains the same as the high-res model described in Section~\ref{sec:ssc}. We mention the differences for the low-res and mid-res models.

\noindent\textbf{Low-Res Model: } The low-res model inputs $270{\times}480$ images. As such, it can have more parameters than a mid-res or hi-res model that inputs $4\times$ and $16\times$ higher resolution inputs respectively. We use a $n_f = 28$ for this model. The input images are zero-padded, and make them $320{\times}512$ in dimensions. 

\noindent\textbf{Mid-Res Model: } We use a $n_f = 23$ for this model. The input images are zero-padded, and make them $640{\times}1024$ in dimensions. 

\begin{figure*}[t]
\centering
\includegraphics[width=1.0\linewidth]{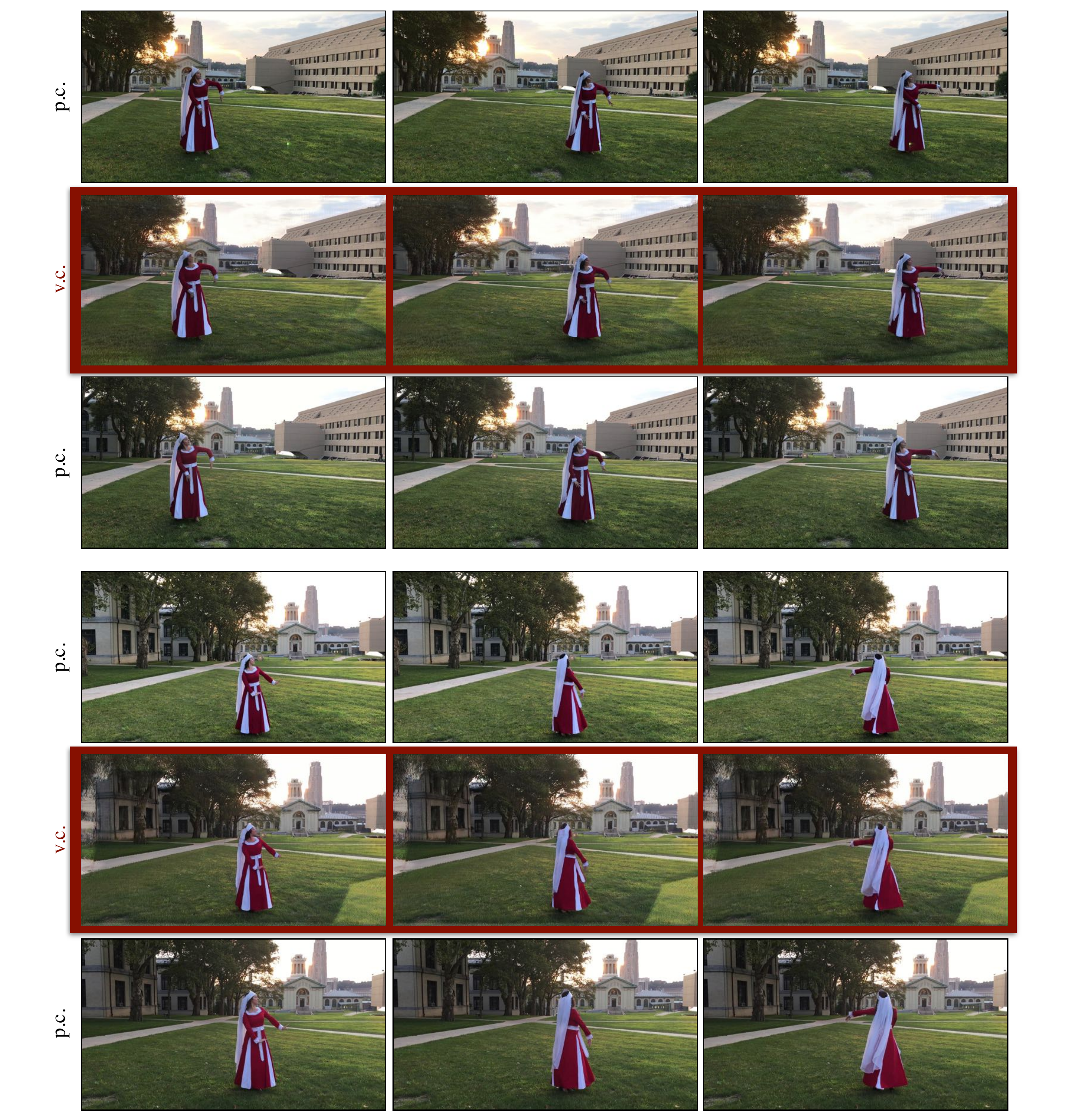}
\caption{\textbf{Flowing Dress and Open Hair: } We show two examples of virtual cameras generated for the Western Folk Dance sequence in which the performer is wearing a flowing dress with open hairs. For each virtual camera (v.c.), we show the physical cameras (p.c.) on its sides. Zoom-in for the detailed human motion and the dress in these two examples.}
\label{fig:dress}
\end{figure*}

\subsection{Limited Data and Overfitting}

\begin{figure}[t]
\centering
\includegraphics[width=1.0\linewidth]{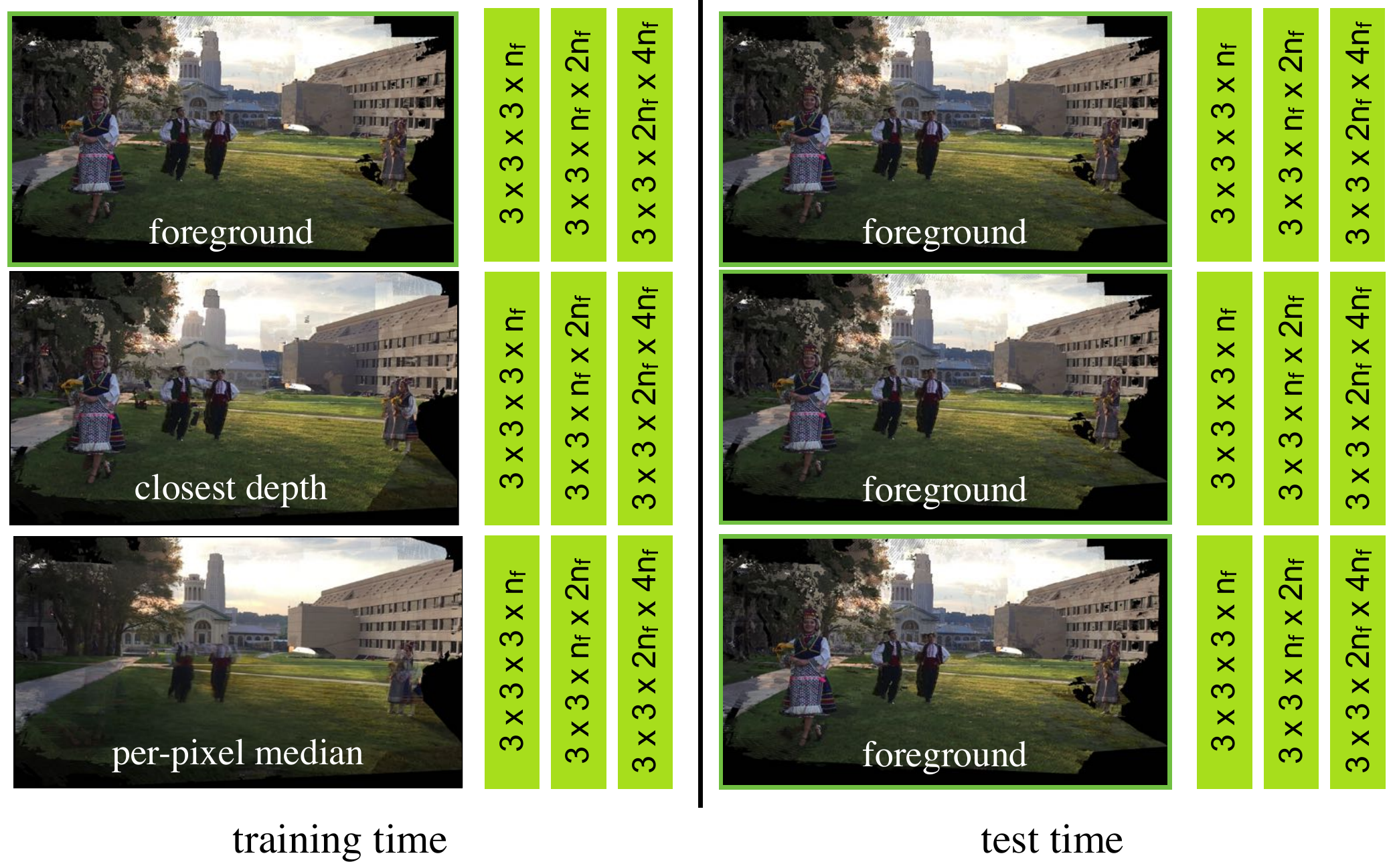}
\caption{\textbf{Foreground Streams: } We use three parallel streams of convolution to encode foreground information. At training time, we use different foreground estimates (see Section~\ref{ssn:fg}) as input to each stream. At test time, we use the same foreground estimated using the consensus depth for all streams. }
\label{fig:4d-network-fg}
\end{figure}

\begin{figure}[t]
\centering
\includegraphics[width=1.0\linewidth]{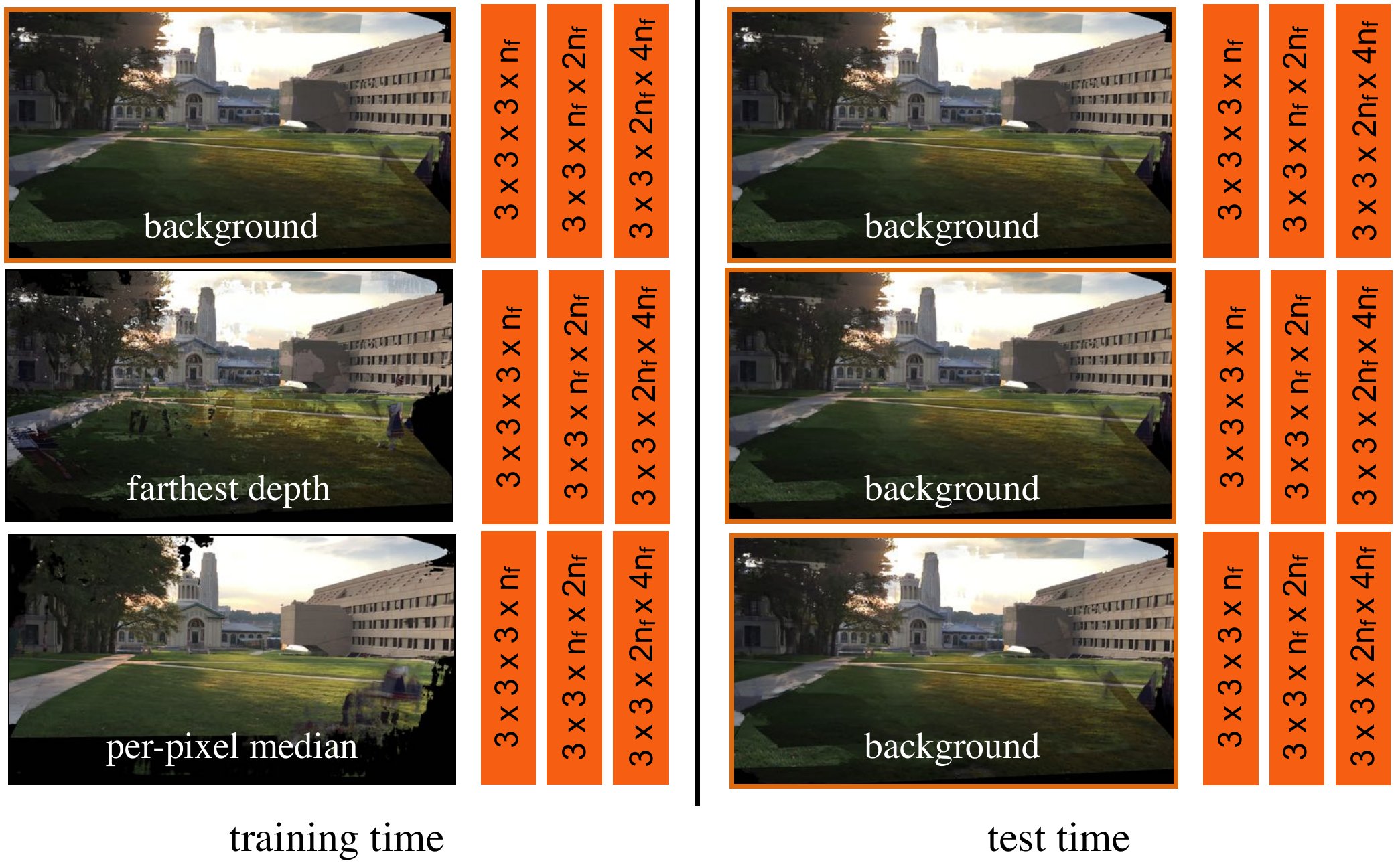}
\caption{\textbf{Background Streams: } We use three parallel streams of convolution to encode background information. At training time, we use different background estimates (see Section~\ref{ssn:bg}) as input to each stream. At test time, we use the same background estimated using our long-term spatiotemporal accumulation method for all streams.}
\label{fig:4d-network-bg}
\end{figure}

In this work, we train a CNN specific to a scene/sequence. Owing to limited data for a sequence, the models are susceptible to overfit to specific inputs and fail to generalize for a wide range of camera views not available at training time. We could not naively use very high capacity models and train them using limited input-output pairs. Now, there are two popular ways to increase the capacity of CNNs: (1) adding more convolutions to make them deeper; and (2) increasing the number of filters in each convolutional layer. We take a hybrid of these methods. We add more convolutions, but in parallel and not sequential, which also increase the number of filters in the later parts of our network. Specifically, we add more foreground and background streams in our network. Figure~\ref{fig:4d-network} shows one stream each of foreground and background feeding to the U-Net. We use three streams each for foreground and background in our setup as shown in Figure~\ref{fig:4d-network-fg} and Figure~\ref{fig:4d-network-bg}. We use different inputs for each stream at training time to provide variability in the data.

\noindent\textbf{Foreground Streams: }  Figure~\ref{fig:4d-network-fg} shows three parallel foreground streams used to encode foreground information. We use different foreground estimates as an input to each stream. We refer the reader to Section~\ref{ssn:fg} for the details of each input shown in Figure~\ref{fig:4d-network-fg}. At test time, we only use our foreground estimated using the consensus depth for all streams.

\noindent\textbf{Background Streams: } Similarly, Figure~\ref{fig:4d-network-bg} shows three parallel background streams used to encode background information. We use different background estimates as an input to each stream. We refer the reader to Section~\ref{ssn:fg} and Section~\ref{ssn:bg} for the details of each input shown in Figure~\ref{fig:4d-network-bg}\footnote{We have not described \emph{per-pixel median} for the background in Section~\ref{ssn:bg}. It refers to the spatiotemporal accumulation done over median foreground images.}. At test time, we only use our static background estimated using our long-term spatiotemporal accumulation method for all streams.

The addition of multiple streams allow us to increase the capacity of the model and use of different inputs provide variability in the data. This enables us to train a model without overfitting to specific input-output pairs.

\begin{figure*}[t]
\centering
\includegraphics[width=1.0\linewidth]{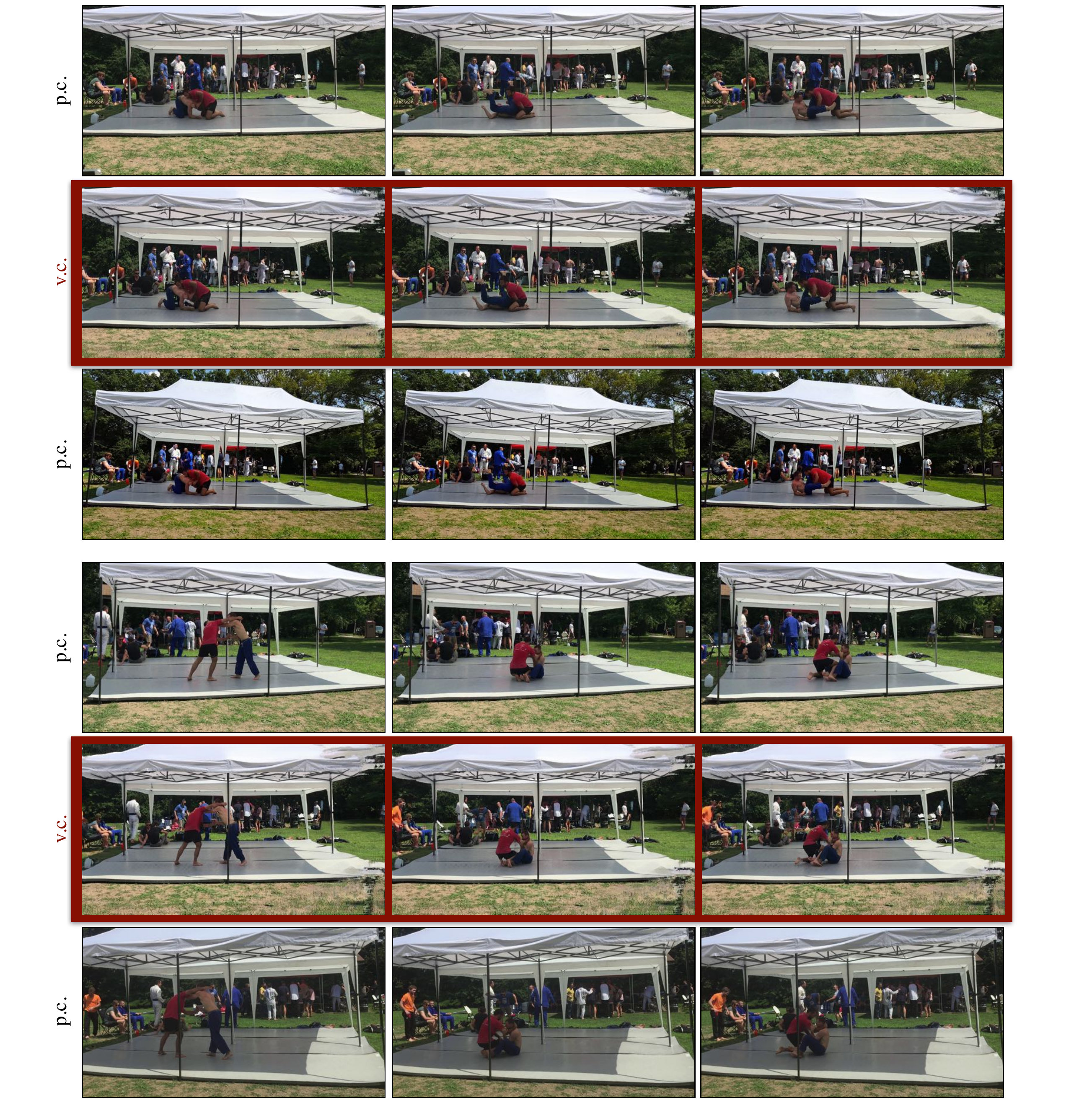}
\caption{\textbf{Many People and Unchoreographed Sequence: } We show two examples of virtual cameras generated for Jiu-Jitsu Retreat sequence that is an example of capture with many people and unchoreographed event. For each virtual camera (v.c.), we show the physical cameras (p.c.) on its sides.}
\label{fig:many_people}
\end{figure*}

%% file: multi-view-sequences.tex
\section{Unconstrained Multi-View Sequences}
\label{sec:mvseq}

 We collected a large number of highly diverse sequences of unrestricted dynamic events consisting of humans and birds. These sequences were captured in different environments and varying activities using up to $15$ hand-held mobile phones. At times, we also mount the cameras on tripod stand as a proxy of hand-held capture. Note that an actual hand-held capture is better than one mounted on tripod for two reasons: (1) diversity of captured views for training our model; (2) the videographer bias enables a better capture of the event. We now describe a few sequences used in this work to get a better sense of the data.

\begin{figure*}[t]
\centering
\includegraphics[width=1.0\linewidth]{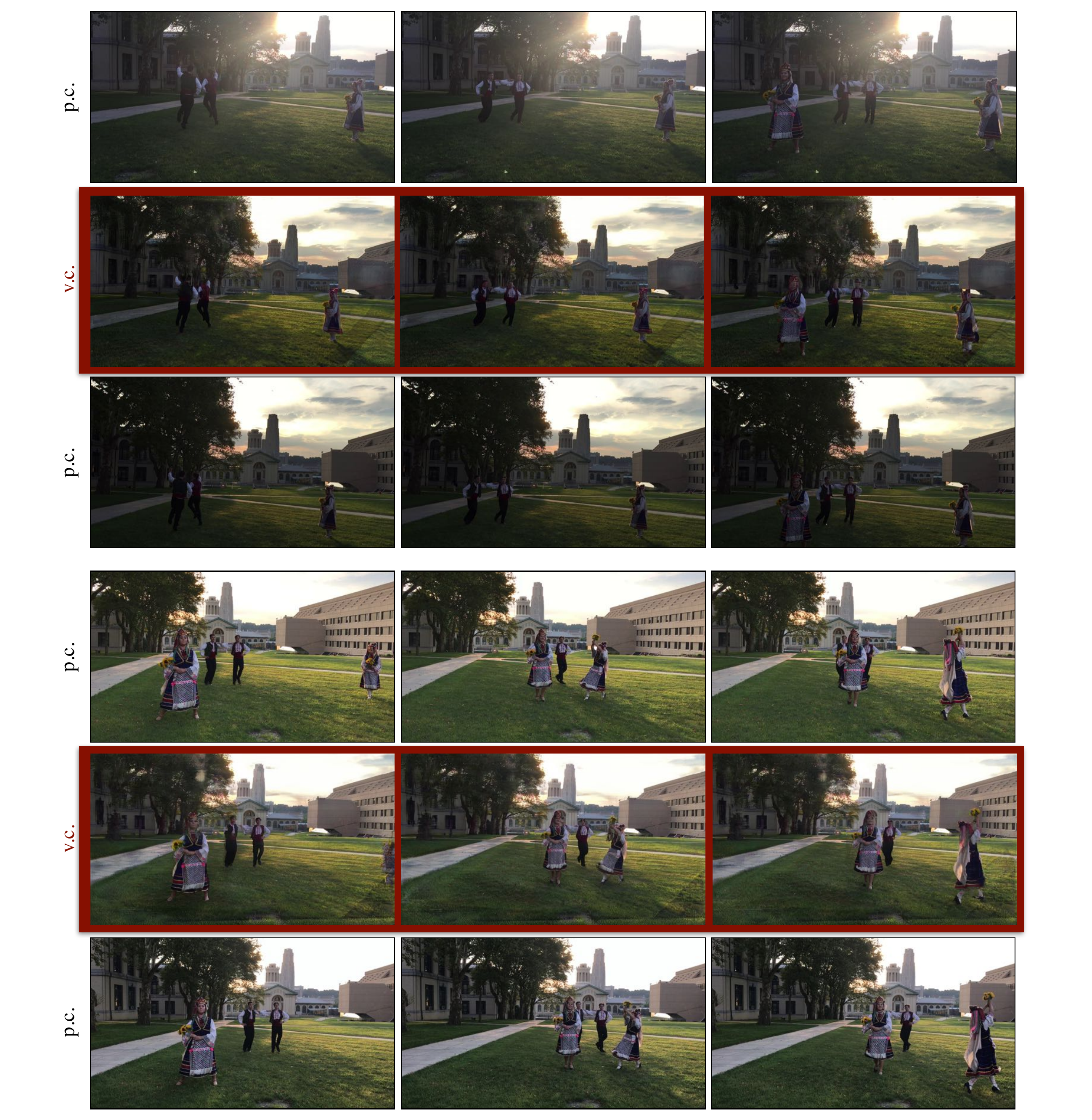}
\caption{\textbf{Challenging Illumination and Multiple Dancers: } We show two examples of virtual cameras generated for another Western Folk Dance sequence with challenging illumination condition and self occlusion due to multiple dancers. For each virtual camera (v.c.), we show the physical cameras (p.c.) on its sides.}
\label{fig:light}
\end{figure*}

\subsection{Human Performances}
\label{sec:mv-humans}
We captured a wide variety of human motion, human-human interaction, human-object interaction, clothing, both indoor and outdoor, under  varying environmental and illumination conditions. 

\noindent\textbf{Western Folk Dance: } We captured sequences of western folk dance performances. This sequence is challenging due to \emph{flowing} dresses worn by performers, open hair, self-occlusions, and illumination conditions. We believe that this sequence also paves the path for incorporating illumination condition in the future work. Figure~\ref{fig:human_dataset}-1 shows one of the two western folk dance sequences that we captured.

We show examples of virtual cameras created using our approach for these sequences in Figure~\ref{fig:dress} and Figure~\ref{fig:light} .

\noindent\textbf{Jiu-Jitsu Retreat: } Jiu-Jitsu is a type of Brazilian Martial Art. We captured sequences of this sporting event during a summer retreat of the Pittsburgh Jiu-Jitsu group. This sequence is an extreme example of unchoreographed dynamic motion from more than $30$ people who participated in it. Figure~\ref{fig:human_dataset}-2 shows the capture of a JiuJitsu event in the foreground with arbitrary human motion in a picnic in the background.

We show examples of virtual camera created for this sequence in Figure~\ref{fig:many_people}.

\noindent\textbf{Tango: } We captured sequences of Tango dance in an indoor environment. Both performers wore proper dress for Tango. Self-occlusion between the performers makes it challenging for 4D visualization. Shown in Figure~\ref{fig:human_dataset}-3 is an example of one of the four Tango dance sequences that we captured. Note the reflections on the semi-glossy ground and featureless surroundings.

\noindent\textbf{Performance Dance: } We captured many short performance dances including Ballet, and reenactments of plays. These sequences were collected inside an auditorium. The lighting condition, clothing, and motion change drastically in these sequences. Figure~\ref{fig:human_dataset}-4 shows an example of performance with an extremely wide baseline and challenging illumination.

\subsection{Bird Sequences}
\label{sec:mv-birds}

We captured a wide variety of birds at the National Aviary of Pittsburgh. We have no control on the motion of birds, their environment, lighting condition, and dynamism in the background due to human movement in the aviary.

\noindent\textbf{Wetlands: } The American Flamingos are the most popular wetlands bird at the National Aviary of Pittsburgh. In this sequence, we captured a free motion of many American Flamingos from multiple cameras. Slowly moving water with reflection of surroundings in which these birds live makes it challenging and appealing  (Figure~\ref{fig:bird_dataset}-1). There may be an occasional sight of Brown Pelicans and Roseate Spoonbills in this sequence.

\noindent\textbf{Penguins: } There are around 20 African Penguins at the Penguin Point in National Aviary of Pittsburgh. Penguin Point consists of rocky terrain and a pool for penguins. The arbitrary motion of penguins and reflection in water makes this sequence totally uncontrolled. Figure~\ref{fig:bird_dataset}-2 shows an example of a sequence captured at the Penguin Point. There is also a frequent movement of humans in this sequence making the background dynamic.

\noindent\textbf{Tropical Rain-forests:} There are a wide variety of tropical rain forest birds at the National Aviary of Pittsburgh. These include Bubba the Palm Cockatoo, a critically endangered parrot species, Gus and Mrs. Gus the Great Argus Pheasants, a flock of Victoria Crowned Pigeons, Southern Bald Ibis, Guam Rails, Laughing Thrushes, Hyacinth Macaws, and a two-toed Sloth. The dense trees in this section act as a natural occlusion while capturing the birds and makes it exciting for 4D capture. Figure~\ref{fig:bird_dataset}-3 shows an example of a sequence that captured Victoria Crowned Pigeons.

%% file: experiments.tex
\section{Quantitative Analysis }
\label{sec:experiments}

We used sequences from Vo et al.~\cite{Vo_2016_CVPR} to properly compare our results with their 3D reconstruction (SfM+humans). Figure~\ref{fig:concerns} and Figure~\ref{fig:overview} shows the results of freezing the time and exploring the views for these sequences.

\begin{table}
\small{
\setlength{\tabcolsep}{3pt}
\def\arraystretch{1.3}
\center
\begin{tabular}{@{}l  c c c c c }
\toprule
\textbf{Approach} &   M.S.E   &  PSNR & SSIM &  LPIPS~\cite{zhang2018perceptual} & FID~\cite{NIPS2017_7240} \\
\midrule
N.N & $5577.65$ & $10.72$ & $0.27$ &$0.57$ &  - \\
(same-time) &$\pm 927.47$ & $\pm 0.73$ & $\pm 0.05$ &$\pm 0.07$ & \\
 \midrule
 N.N & $4948.64$ & $11.29$ & $0.31$ &$0.50$ &  - \\
(all-time) &$\pm 1032.76$ & $\pm 1.00$ & $\pm 0.04$ &$\pm 0.11$ & \\
\midrule
SfM & $3982.26$   & $12.25$ & $0.33$ & $0.45$ & $190.84$ \\
+ Humans&$\pm 1050.28$ & $\pm 1.02$ & $\pm 0.08$ & $\pm 0.022$ & \\
\midrule
Inst. & $5956.25$ &  $11.18$& $0.42$ & $0.43$ & $100.10$\\
&$\pm 3139.61$ & $\pm 2.89$ & $\pm 0.17$ & $\pm 0.16$ & \\
\midrule
Ours & \bm{$714.95$} & \bm{$20.14$} & \bm{$0.79$}  & \bm{$0.13$} & \bm{$21.98$} \\
& \bm{$\pm 364.99$} &\bm{ $\pm 2.21$} & \bm{$\pm 0.07$} & \bm{$\pm 0.06$} & \\
\bottomrule
\end{tabular}
\vspace{0.2cm}
\caption{\textbf{Comparison: } We contrast our approach with: (1). a simple nearest neighbor (\textbf{N.N.}) baseline. There are two kinds of nearest neighbors in our scenario. The first is the nearest camera view at the \textbf{same time} instant and the other is the nearest camera view across \textbf{all time} instants. (2). reconstructed outputs of \textbf{SfM+humans}; and finally (3). instantaneous foreground (\textbf{Inst}) defined in Section~\ref{ssn:fg}. We use various evaluation crietria to study our approach in comparisons with these three methods: (1). \textbf{M.S.E}: We compute a mean-squared error of the generated camera sequences using held-out camera sequences.; (2). \textbf{PSNR}: We compute a peak signal-to-noise ratio of the generated sequences against the held out sequences; (3). \textbf{SSIM}: We also compute a SSIM in similar manner.; (4). We also use LPIPS~\cite{zhang2018perceptual} to study structural similarity and to avoid any biases due to MSE, PSNR, and SSIM. Lower it is, better it is. Note that all the above four criteria are computed using held-out camera sequences; and finally (5) we compute a \textbf{FID}-score~\cite{NIPS2017_7240} to study the quality of generations when a ground-truth is not available for comparisons. Lower it is, better it is.}
\label{tab:obj}
}
\end{table}

\noindent\textbf{Evaluation: } We use a mean-squared error (MSE), PSNR, SSIM, and LPIPS~\cite{zhang2018perceptual} to study the quality of virtual camera views created using our approach. \textbf{MSE}: Lower is better. \textbf{PSNR}: Higher is better. \textbf{SSIM}: Higher is better. \textbf{LPIPS}: Lower is better. We use held-out cameras for proper evaluation. We also compute a \textbf{FID} score~\cite{NIPS2017_7240}, lower the better, to study the quality of sequences where we do not have any ground truth (e.g., freezing the time and exploring views). This criterion contrast the distribution of virtual cameras against the physical cameras.\\

\textbf{Baselines:} To the best of our knowledge, there does not exist a work that has demonstrated dense 4D visualization for in-the-wild dynamic events captured from unconstrained multi-view videos. We, however, study the performance of our approach with: (1) a simple nearest neighbor baseline \textbf{N.N.}: We find nearest neighbors of generated sequences using conv-5 features of an ImageNet pre-trained AlexNet model. This feature space helps in finding the images closer in structure. Additionally, there are two kinds of nearest neighbors in our scenario. The first is the nearest camera view at the \textbf{same time} instant and the other is the nearest camera view across \textbf{all time} instants.; (2) \textbf{SfM+humans}: We use work from Vo et al~\cite{Vo_2016_CVPR,vo2018automatic} for these results.; and finally (3) we contrast it with instantenous foreground image (\textbf{Inst}) that is defined in Section~\ref{ssn:fg}.

Table~\ref{tab:obj} contrasts our approach with various baselines on held-out cameras for different sequences. We observe significantly better outputs under all the criteria. We provide more qualitative analysis on our project page -- \url{http://www.cs.cmu.edu/~aayushb/Open4D/}

\begin{figure*}[t]
\centering
\includegraphics[width=1.0\linewidth]{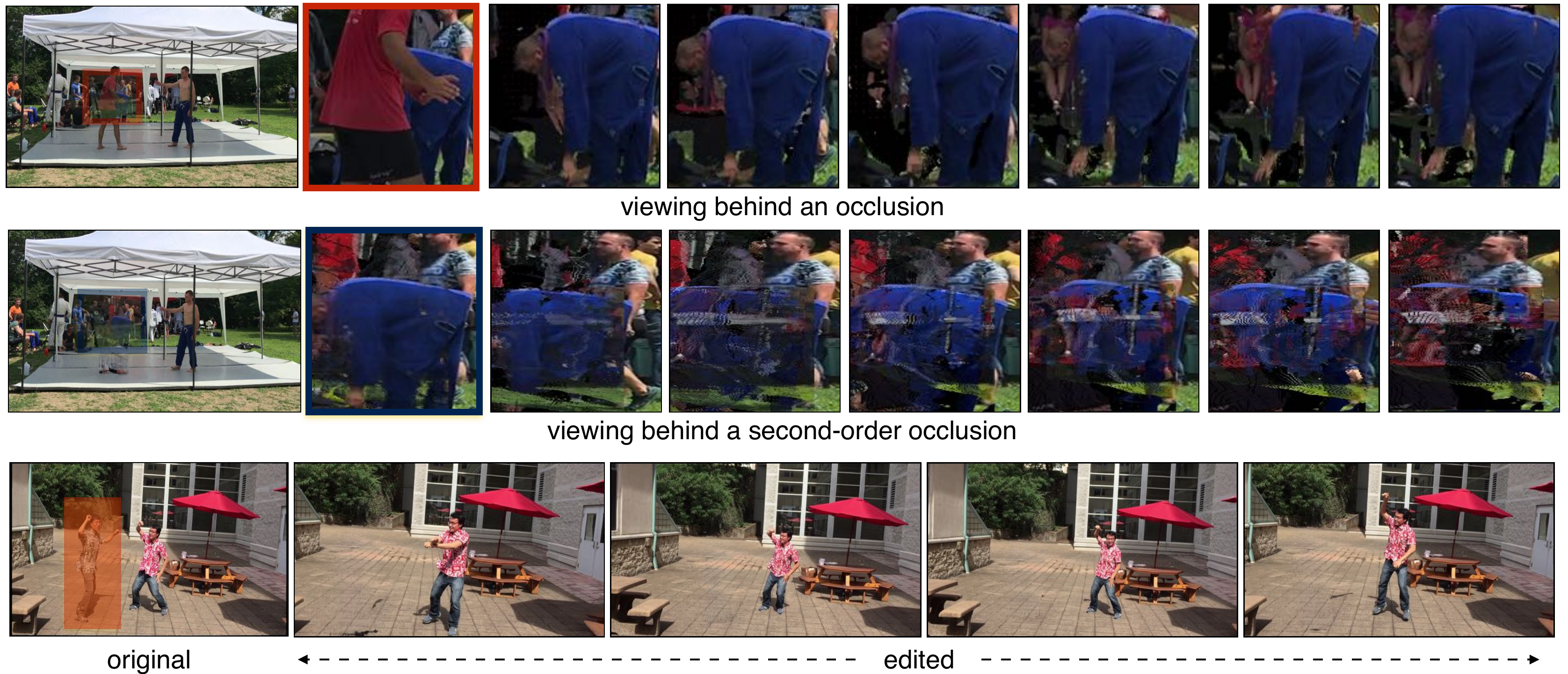}
\caption{\textbf{User-Controlled Editing: } We show two examples of user-controlled editing in videos. In the \textbf{top}-row, a user selects a mask to see the occluded blue-shirt person (behind red shirt person). There is no way we can infer this information from a single-view. However, multi view information allows us to not only see the occluded human but also gives a sense of activity he is doing. We show frames from 2 seconds of video. In the \textbf{middle}-row, we want to see the part of scene behind the blue-shirt person who is \emph{disoccluded} above. This is an example of a seeing behind a second-order occlusion. While not as  sharp as first-order occlusion result, we can still see green grass and white bench in the background with a person moving. This particular scenario is not only challenging due to second-order occlusion but also because of larger distance from cameras. In the \textbf{bottom}-row, a user can remove the foreground person by marking on a single frame in video. Our system associates this mask to all the frames in video, and edit it to show background in place of human. We show frames of edited video (20 seconds long).} 
\label{fig:manipulation}
\end{figure*}

\section{User-Controlled Editing}

We have complete control of the 3D space and time information of the event. This 4D control allows us to browse the dynamic events. A user can see behind the occlusions if a certain information is visible in other views. A user can also edit, add, or remove objects. To accomplish this, a user marks the required portion in a video. Our approach automatically edits the content, i.e., update the background and foreground, via multi-view information — the modified inputs to stacked multi-stage composition results in desirable outputs. Importantly, marking on a single frame in the video is sufficient, as we can  propagate the mask to the rest of the video (4D control of foreground).  We show two examples of user-controlled editing in Figure~\ref{fig:manipulation}. In the first example, we enable a user to see occluded person without changing the view. Our system takes input of mask from the user, and \emph{disocclude} the blue-shirt person (Figure~\ref{fig:manipulation}-top-row). We also explore viewing behind a second-order occlusion.  Figure~\ref{fig:manipulation}-middle shows a very challenging example of viewing behind the blue-shirt person. Despite farther away from the camera, we see grass, white table, and a person moving in the output. Finally, we show an example of editing where a user can mark region in a frame of video (Figure~\ref{fig:manipulation}-bottom-row). Our system generates full video sequence without the masked person.

%% file: discussion.tex
\section{Discussion \& Future Work}
\label{sec:discussion}

The world is our studio. The ability to do 4D visualization of dynamic events captured from unconstrained multi-view videos opens up avenue for future research to capture events with a combination of drones, robots, and hand-held cameras. The use of self-supervised and scene-specific CNNs allows one to browse the 4D space-time of dynamic events captured from unconstrained multi-view videos.  We extensively captured various in-the-wild events to study this problem. We show different qualitative and quantitative analysis in our study. A real-time user guided system that allows a user to upload videos and browse will enable a better understanding of 4D visualization systems. The proposed formulation and the captured sequences, however, open a number of opportunities for future research such as incorporating illumination and shadows in 4D spatiotemporal representation, and modeling low-level high frequency details. One drawback of our method is that the video streams are treated as perfectly synchronized. This introduces motion artifacts for fast actions~\cite{Vo_2016_CVPR}. Future work will incorporate sub-frame modeling between different video streams in depth estimation and view synthesis modules for more appealing 4D slow motion browsing.

\vspace{.5cm}
{\noindent\textbf{Acknowledgements: }  We are extremely grateful to Bojan Vrcelj for helping us shape the project. We are also thankful to Gengshan Yang for his help with the disparity estimation code and many other friends for their patience in collecting the various sequences. We list them on our project page. This work is supported by the Qualcomm Innovation Fellowship, NSF CNS-1446601 and ONR N00014-14-1-0595 grant.}